# Enhancing Decision Analysis with a Large Language Model: pyDecision a Comprehensive Library of MCDA Methods in Python

**Valdecy Pereira**

**(Corresponding Author)**

**valdecy.pereira@gmail.com**

**Department of Production Engineering, Federal Fluminense University, Rua Passo da Pátria, CEP: 24210-240, São Domingos, Niterói, RJ, Brazil. ORCID: 0000-0003-0599-8888**

**Marcio Pereira Basilio**

**marciopbasilio@gmail.com**

**Department of Production Engineering, Federal Fluminense University, Brazil. ORCID: 0000-0002-9453-741X**

**Carlos Henrique Tarjano Santos**

**tesserato@hotmail.com**

**Department of Production Engineering, Federal Fluminense University, Brazil. ORCID: 0000-0003-3134-2778**

**Abstract**

*Purpose*: Multicriteria decision analysis (MCDA) has become increasingly essential for decision-making in complex environments. In response to this need, the pyDecision library, implemented in Python and available at https://bit.ly/3tLFGtH, has been developed to provide a comprehensive and accessible collection of MCDA methods. *Methods*: The pyDecision offers 70 MCDA methods, including AHP, TOPSIS, and the PROMETHEE and ELECTRE families. Beyond offering a vast range of techniques, the library provides visualization tools for more intuitive results interpretation. In addition to these features, pyDecision has integrated ChatGPT, an advanced Large Language Model, where decision-makers can use ChatGPT to discuss and compare the outcomes of different methods, providing a more interactive and intuitive understanding of the solutions. *Findings*: Large Language Models are undeniably potent but can sometimes be a double-edged sword. Its answers may be misleading without rigorous verification of its outputs, especially for researchers lacking deep domain expertise. It's imperative to approach its insights with a discerning eye and a solid foundation in the relevant field. *Originality*: With the integration of MCDA methods and ChatGPT, pyDecision is a significant contribution to the scientific community, as it is an invaluable resource for researchers, practitioners, and decision-makers navigating complex decision-making problems and seeking the most appropriate solutions based on MCDA methods.

**Keywords**: Multicriteria Decision Analysis; Python; Decision Making

## 1. Introduction

Multicriteria Decision Analysis (MCDA) is a decision-making methodology widely used in many fields, such as finance, engineering, environmental science, and public policy. It is instrumental when decision-makers must make complex decisions based on multiple criteria, so MCDA methods can help decision-makers identify the best alternative among feasible options. However, the increasing complexity of decision-making problems across various domains has increased the demand for robust, easy-to-use tools (TAHERDOOST & MADANCHIAN, 2023; PEREIRA et al., 2017).

Cinelli et al. (2021) emphasize the escalating significance of MCDA techniques in decision-making, which has consequently fostered the emergence of numerous software tools and libraries encapsulating these methods. Primarily, these tools contribute to formulating the decision issue, enabling decision-makers to systematically and consistently discern relevant criteria and alternatives. Afterward, they simplify the process of appraising alternatives through various techniques. These tools foster the conveyance and visualization of outcomes, equipping decision-makers with coherent and concise data to fortify their decision-making efforts.

Cinelli et al. (2014) highlight the rich ecosystem of software tools available in the MCDA sphere. This varied collection includes traditional options such as Expert Choice (Expert Choice, 2020), which embodies the Analytic Hierarchy Process (AHP) methodology, and SuperDecisions (CREATIVE DECISIONS FOUNDATION, 2020), a tool designed to support AHP and the Analytic Network Process (ANP) methodologies. The breadth and diversity of these software offerings underscore the extensive applications of MCDA and the ongoing evolution of decision-making methodologies.

Despite their pervasive utilization, MCDA software tools are not devoid of limitations, resulting in several unaddressed gaps. For example, software such as Expert Choice concentrates solely on a specific method like AHP, constraining users who seek to explore and compare various MCDA techniques for their decision-making challenges. The learning curve associated with some MCDA software tools can be pronounced, necessitating significant time and effort from users to comprehend and master the software's intricacies and capabilities. Moreover, commercial MCDA software tools may carry substantial costs, creating barriers for smaller organizations or individual researchers, and the proprietary nature of closed-source MCDA software can impede collaboration and ongoing refinement. These gaps and challenges underscore the imperative for the continuous advancement and betterment of MCDA software tools, ensuring they effectively cater to the ever-evolving needs of researchers and practitioners.

As mentioned above, an ideal MCDA software should address a variety of considerations. It should support multiple MCDA methods, enabling users to explore and compare diverse techniques while offering customization to adapt the decision-making process to their needs. User-friendliness, intuitive interface, and comprehensive documentation should reduce the learning curve. As an open-source platform, the software would foster collaboration and continuous improvement, allowing users to stay abreast of the latest advancements. Seamless integration with other data management, analysis, or visualization tools would simplify incorporating MCDA results into broader workflows.

Large Language Models (LLM) have significantly advanced Natural Language Processing (NLP); these models can generate human-like text, understand context, and respond to queries in a highly interactive and engaging manner (KIM et al., 2023). This capability can be integrated into MCDA, revolutionizing how complex decision-making problems are approached. Integrating LLM enables a conversational interface for the decision-making process, which is particularly helpful in explaining the relative weights of different criteria, the outranking of various alternatives, and how final decisions are derived. This approach offers a highly interactive and intuitive way to navigate the intricacies of MCDA, making complex decision-making more accessible and understandable.

The marriage of MCDA software functionality and the cutting-edge capabilities of LLM forms a revolutionary approach that streamlines and enhances complex decision-making processes. This innovative amalgamation facilitates an interactive, user-focused platform that bridges the gap between the computational rigidity of MCDA and the dynamic conversational nature of LLM, enlivening the decision-making process in an illuminating and engaging manner. Furthermore, Python emerges as an ideal environment for executing this integration. Renowned for its versatility, user-friendliness, and robust open-source community, Python's simplicity, coupled with a vast selection of libraries, allows for smooth integration with a wide array of data management, analysis, and visualization tools.

In the context of the current landscape of MCDA software and advancements in LLMs, we introduce pyDecision, a comprehensive and pioneering Python library poised to transform the MCDA domain. Our library equips its users with the capability to tailor an impressive array of 70 distinct MCDA methods to their unique decision-making needs, including widely acclaimed methods such as AHP, TOPSIS (Technique for Order of Preference by Similarity to Ideal Solution), and ELECTRE (ELimination Et Choix Traduisant laRÉalité) family, among others. At the heart of pyDecision lies a comprehensive feature suite that empowers users to construct, assess, and unravel complex decision-making scenarios with unparalleled depth and clarity.

This article comprehensively explores the pyDecision library, encapsulating its extensive capabilities and unique offerings. Our narrative initiates with an introduction to the landscape of MCDA software tools, including the implemented MCDA methods in pyDecision, highlighted in Section 2. Section 3 explores the pyDecision library, detailing its capabilities to compare rank generation methods and integration with a Large Language Model, underscoring the synergy of this combination. Section 4 compares methods that can generate weights and their integration with a Large Language Model. Conclusively, Section 5 encapsulates the indelible contributions pyDecision has made to the scientific community.

## 2. MCDA Softwares

MCDA, as a structured methodology, enhances decision-making processes by guiding the identification of potential alternatives, the selection of assessment criteria, and the consolidation of various stakeholder preferences. This methodology's robustness and utility have been recognized across diverse fields, leading to a substantial increase in dedicated software tools designed to streamline and optimize MCDA processes. (BISDORF et al., 2015; GRECO et al., 2016).

In this context, we adopt a more expansive definition of "software," encompassing not only standalone applications but also various libraries and packages. Libraries and packages are collections of pre-compiled routines, functions, and multiple resources. Libraries are typically organized sets of reusable code and functions that developers can call upon within their programs, enhancing efficiency and reducing redundancy. Packages are similar in function to libraries but often include additional components like documentation, configuration data, and other ancillary resources. Both libraries and packages provide a way to encapsulate proven solutions, promoting reusability, maintainability, and the collaborative evolution of software (SOMMERVILLE, 2011).

MCDA softwares can be systematically divided into four distinct classes based on the breadth of model implementation and the accessibility of source code. The first class, Single/Family Model & Proprietary, consists of tools that narrowly focus on one specific model or a closely related family of models and are proprietary, limiting adaptability and imposing constraints on customization and transparency. The second class, Single/Family Model & Open Source, includes tools that, while similarly focusing on a singular model or related family of models, are open source, promoting transparency and customization but potentially restricted in versatility. The third class, Multi-Model & Proprietary, encapsulates software that offers a broader range of models but is proprietary, hindering customizability, transparency, and community collaboration. Lastly, the fourth class, Multi-Model & Open Source, comprises open source tools and supports multiple models, fostering broad methodological flexibility and adaptability and encouraging community-driven enhancements. Condensing and illustrating:

a) Single/Family Model & Proprietary

1000minds (1000MINDS, 2023), a web-based application, epitomizes this approach by exclusively utilizing the PAPRIKA (Potentially All Pairwise RanKings of all possible Alternatives) method, created by Hansen & Ombler (2008) to cater to decision-making processes. As a proprietary platform, 1000minds mitigates some limitations by providing complimentary access to academic researchers and students, enhancing accessibility within certain academic and research communities. Another example is Expert Choice, a proprietary tool and a specialized decision-making software operationalizing the AHP method.

b) Single/Family Model & Open Source

VoraciousAHP (PEREIRA, 2022), 3MO-AHP (FLORIANO et al., 2022), SuperDecisions, J-Electre (PEREIRA & NEPOMUCENO, 2021), and Electre Tree (MONTENEGRO DE BARROS et al., 2021) collectively represent a range of solutions within the MCDA domain that implement single or family models and are open source. VaraciousAHP employs the AHP method and can refine the Pairwise Comparison Matrix for consistency. This notion of consistency in AHP is also addressed by 3MO-AHP, which extends its utility to both traditional AHP and Fuzzy AHP, giving multiple solutions depending on the decision-maker requirements. As a freely available tool, Super Decisions is designed to facilitate the learning and implementation of the AHP and ANP methodologies. In contrast, J-Electre offers a dedicated platform implementing the ELECTRE family of methods, including ELECTRE I, Is, Iv, II, III, IV, and Tri, providing versatility across various MCDA scenarios. Finally, Electre Tree, which specializes in the Electre Tri algorithm, innovates by inferring missing parameters and accommodating incomplete information without sacrificing robustness.

c) Multi-Model & Proprietary

Criterium DecisionPlus (INFOHARVEST, 2023) is an MCDA-based software for decision-making processes. It prominently implements the AHP and the Simple Multi-Attribute Rating Technique (SMART). D-Sight (ENTERPRISE SOFTWARE, 2023) focuses on decision support software and services. Specifically, D-Sight's tools utilize the Preference Ranking Organization Method for Enrichment Evaluation (PROMETHEE), Geometrical Analysis for Interactive Decision Aid (GAIA), Multi-Attribute Utility Theory (MAUT), and AHP.

d) Multi-Model & Open Source

In the dynamic field of MCDA, a plethora of libraries and packages have been developed, each boasting unique language foundations and functionalities. Ranking Trees (RODRIGUES et al., 2021) offers a specialized algorithm capable of eliciting various combinations of parameters for ELECTRE II, II, and IV and PROMETHEE I, II, III, and IV methods. The Python libraries pyFDM (WIĘCKOWSKI et al., 2023a) and pyIFDM (WIĘCKOWSKI et al., 2023b) provide sophisticated tools for handling uncertain data and intuitionistic fuzzy MCDA methods, respectively. Further extending Python's reach in this area, pymcdm (KIZIELEWICZ et al., 2023) and pyrepo-mcda (WĄTRÓBSKI et al., 2023) encompass a broad array of MCDA methods. Meanwhile, R MCDA (BIGARET et al., 2023) capitalizes on the statistical prowess of the R environment to deploy various algorithms, and JMcDM (SATMAN et al., 2021) introduces the Julia language to multiple-criteria decision-making with an intuitively designed interface. These open-source libraries, featuring various MCDA methods, contribute to a rich and multifaceted decision-making ecosystem.

In this study, we believe that the Multi-Model & Open Source category represents an optimal choice because it offers a blend of methodological flexibility, customization options, and community-driven enhancements. Unlike proprietary models, open-source platforms enable complete transparency, allowing for rigorous academic scrutiny and ongoing improvements. These platforms often boast a wide range of MCDA methods, thus providing comprehensive solutions suitable for diverse decision-making scenarios. Furthermore, being open-source means that these tools can be customized to meet the unique requirements of any specific project or research endeavor. This adaptability encourages an inclusive, collaborative environment where methods and tools are continually refined, extended, and updated by a global community of researchers and practitioners.

Therefore, in light of the advantages mentioned earlier, libraries in the Multi-Model & Open Source category offer a rich tapestry of methodologies, enabling a more robust and adaptable framework for multiple-criteria decision-making. A detailed comparison of these libraries, juxtaposed with pyDecision, which also follows a Multi-Model & Open Source approach, is available in Table 1. In this comparison, an (X) denotes the implementation of the classic method, (F) signifies the implementation of the fuzzy version, and (I) represents the incorporation of the intuitionistic fuzzy version. Also, the methods marked with an asterisk (∗) indicate that they only generate weights for each criterion, and the hashtag (#) indicates voting methods adapted to the MCDA context.

**Table 1 - MCDA Methods**

| | AHP (Saaty, 1990) | ARAS (Zavadskas & Turskis, 2010) | #Borda (Borda, 1781) | *BWM (Rezael, 2015) | *CILOS (Zavadskas, & Podvezko, 2016) | CoCoSo (Yazdani et al., 2019) | CODAS (Ghorabaee et al., 2016) | COMET (Sałabun, 2015) | #Copeland (Copeland, 1951) | COPRAS (Zavadskas et al., 1994) | CRADIS (Puška et al., 2022) | *CRITIC (Diakoulaki et al., 1995) | DEMATEL (Fontela, & Gabus, 1976) | EDAS (Ghorabaee et al., 2015) | ELECTRE (Roy, 1968) | *Entropy (Shannon, 1948) | GRA (Deng, 1989) |
|---|---|---|---|---|---|---|---|---|---|---|---|---|---|---|---|---|---|
| pyDecision | XF | XF | X | XF | X | X | X | | X | XF | X | X | XF | XF | X | X | X |
| JMcDM | X | X | | | | X | X | | X | X | | X | X | X | X | X | X |
| pyFDM | | F | | | | | F | | | F | | | | F | | | |
| pyIFDM | | I | | | | | I | | | I | | | | I | | | |
| pymcdm | | X | | | | X | X | X | | X | | | | X | | | |
| pyrepo-mcda | X | X | | | | X | X | | | X | X | | | X | | | |
| R MCDA | X | | | | | | | | | | | | | | X | | |

| | *IDOCRIW (Zavadskas, & Podvezko, 2016) | MABAC (Pamučar & Ćirović, 2015) | MACBETH (Costa & Vansnick, 1994) | MAIRCA (Pamučar et al., 2018) | MARCOS (Stević et al., 2020) | MARE (Hodgett et al.; 2014) | MAUT (Keeney & Raiffa, 1976) | *MEREC (Keshavarz-Ghorabaee et al., 2021) | MOORA (Brauers & Zavadskas, 2006) | MOOSRA (Jagadish & Ray, 2014) | MRSORT (Bouyssou & Marchant, 2007) | MULTIMOORA (Brauers & Zavadskas, 2010) | OCRA (Miloš et al., 2015) | ORESTE (Roubens, 1982) | PIV (Mufazzal & Muzakkir, 2018) | PIROMETHEE (Brans & Vincke, 1985) | PROSA-C (Ziemba, 2020) |
|---|---|---|---|---|---|---|---|---|---|---|---|---|---|---|---|---|---|
| pyDecision | X | X | X | X | X | | X | X | XF | X | | X | XF | X | X | X | |
| JMcDM | | X | | X | X | | | X | X | | | | | | X | X | |
| pyFDM | | F | | F | | | | | F | | | | F | | | | |
| pyIFDM | | I | | I | | | | | I | | | | | | | | |
| pymcdm | | X | | X | X | | | | X | | | | X | | | X | |
| pyrepo-mcda | | X | | | X | | | | | | | X | | | | X | X |
| R MCDA | | | | | | X | | | | | X | | | | | X | |

| | PSI (Maniya & Bhatt, 2010) | Regime (Hinloopen & Nijkamp, 1990) | ROV (Madić et al., 2016) | SAW (Churchman et al., 1954) | SECA (Keshavarz-Ghorabaee et al., 2018) | SMART (Edwards, 1971) | SPOTIS (Dezert et al., 2020) | SRMP (Khannoussi et al.. 2022) | SURE (Hodgett & Siraj, 2019) | TODIM (Gomes & Rangel, 2009) | TOPSIS (Hwang & Yoon, 1981) | UTA (Jacquet-Lagreze & Siskos, 1982) | VIKOR (Opricović, 1990) | WINGS (Michnik, 2013) | WSM (Fishburn, 1967) | WPM (Miller & Starr, 1960) | WASPAS (Zavadskas et al., 2014) |
|---|---|---|---|---|---|---|---|---|---|---|---|---|---|---|---|---|---|
| pyDecision | X | X | X | X | | X | X | | | X | XF | | XF | X | XF | XF | XF |
| JMcDM | X | | X | X | X | | | | | | X | | X | | | X | X |
| pyFDM | | | | | | | | | | | F | | F | | | | |
| pyIFDM | | | | | | | | | | | I | | I | | | | |
| pymcdm | | | | | | | X | | | | X | | X | | | | |
| pyrepo-mcda | | | | X | | | X | | | | X | | X | | | | X |
| R MCDA | | | | | | | | X | X | | X | X | | | | | |

Table 1 serves as a comprehensive guide to the landscape of MCDA libraries, highlighting their specialized methodologies through a structured coding system. Several libraries have emerged to support a variety of methodological approaches. Classic MCDA methods like AHP, ARAS, and SAW remain universally popular across libraries. This universal adoption serves multiple purposes: it offers a reliable foundation for decision-making, acts as an accessible entry point for those new to the field, and provides a rigorous benchmark against which to measure emerging methods. Fuzzy adaptations of these conventional methods are also included, such as Fuzzy ARAS (FU et al., 2021), Fuzzy COPRAS (NARANG et al., 2021), and Fuzzy EDAS (ZINDANI et al., 2019). These fuzzy variants extend the applicability of MCDA techniques to situations demanding more nuanced treatment of uncertainty, thereby broadening the decision-making toolkit available to practitioners.

Weight generation methods like BWM (including its simplified version – Amiri et al.(2021)), CILOS, and CRITIC are crucial, as they set the stage for more complex evaluations. Libraries progressively acknowledge the importance of these weight assignments, underscoring their fundamental role in the accuracy of the overall analysis. Finally, including voting methods such as Borda and Copeland marks an interdisciplinary approach to MCDA, integrating techniques from social choice theory to enrich the decision-making frameworks. Libraries like pyDecision and JMcDM have embraced these methods, reflecting the adaptability of MCDA to incorporate various strategies from different disciplines for more robust decision support. Collectively, these trends signal an MCDA community committed to methodological diversity, foundational rigor, and adaptive innovation to meet the complexities of real-world decision-making.

pyDecision is a comprehensive and adaptable tool within the realm of MCDA. Not only does it support an extensive 70 different MCDA methods, but it also spans a wide gamut from established classics like AHP, ELECTRE (I, I-s, I-v, II, III, IV, and Tri), and PROMETHEE (I, II, III, IV, V, VI, and GAIA), to innovative, modern approaches such as CRADIS and EC PROMETHEE (BASILIO et al. 2023). This diversity makes it versatile enough to address a broad array of decision-making challenges. Furthermore, pyDecision extends its reach by offering fuzzy logic versions of some methods, making it a go-to option for scenarios demanding a more nuanced treatment of uncertainty and ambiguity. It also incorporates advanced criterion-weighting methods like BWM and CRITIC and voting algorithms like Borda and Copeland, providing a rich toolkit for complex decision-making processes. What amplifies pyDecision's utility is its integration of a Large Language Model, specifically ChatGPT (OPENAI, 2023), which allows users to compare results - such as ranks, fuzzy ranks, and weights - generated by various methods. This feature enhances the platform's interactivity and simplifies the intricacies involved in multicriteria decisions. With its broad methodological scope and sophisticated features like LLM, pyDecision tries to serve as a one-stop-shop for individuals and organizations confronting diverse decision-making tasks.

JMcDM emerges as a solid contender focusing on classic and well-established methodologies. It excels in its coverage of time-tested algorithms, including but not limited to AHP, ARAS, ELECTRE I, and PROMETHEE II, further underscoring its credentials as a versatile tool for classical decision-making scenarios. The library also offers the added benefit of criterion weighting techniques like CRITIC and Entropy, thus enhancing its flexibility. Notably, however, it does not support fuzzy methods, which limits its utility in scenarios demanding a nuanced approach to uncertainty or ambiguity. JMcDM is a robust choice for those who prioritize classical, well-tested methods.

pyFDM and pyIFDM distinguish themselves through a laser-focused commitment to fuzzy and intuitionistic fuzzy methods, respectively, providing nuanced alternatives to classic techniques like ARAS and EDAS. Both libraries are invaluable for complex decision-making scenarios that require intricate handling of uncertainty, vagueness, or ambiguity. pyFDM is particularly adept at tackling problems where the limitations of conventional crisp logic become apparent, offering a more nuanced fuzzy approach to managing uncertainty. Conversely, pyIFDM goes further into the specialized realm of intuitionistic fuzzy methods, offering a sophisticated framework for grappling with even more complex forms of uncertainty and ambiguity. While this high level of specialization equips them to excel in their respective niches, it also restricts their applicability to more generalized or classic MCDA tasks. Thus, pyFDM and pyIFDM serve as potent, specialized tools for specific complex scenarios. Still, their narrow methodological focus may not cater to those needing a more comprehensive suite of decision-making instruments.

pymcdm carves out a distinct space as a middle-of-the-road option emphasizing time-tested, classic methods like PROMETHEE II and its unique offering, the COMET technique. This focused commitment to classic methodologies serves as both its strength and limitation. While it facilitates ease of use and offers a simplified user experience, it also constrains the library's versatility in handling more MCDA problems.

pyrepo-mcda emerges as a balanced, mid-tier contender. What distinguishes it is a well-curated selection of classic methods, such as PROMETHEE II, punctuated by specialized offerings like PROSA-C. These unique inclusions give pyrepo-mcda a distinct edge for those interested in exploring beyond the usual staples. Nevertheless, the library has limitations, particularly its lack of support for fuzzy techniques, which narrows its applicability for complex scenarios demanding nuanced handling of uncertainty or vagueness.

R MCDA offers specialized algorithms such as SRMP, SURE, and MARE. R MCDA may not provide an expansive palette of MCDA techniques. Nevertheless, its seamless integration with the R ecosystem and its discerning focus on a handful of reputable methods positions R MCDA as an attractive choice for users grappling with specialized decision-making challenges.

The landscape of MCDA libraries is rich and varied, each distinguished by its strengths, limitations, and specializations. The selection caters to different types of users and problem-specific requirements, ranging from the comprehensiveness of pyDecision to the niche focus of libraries like pyFDM and pyIFDM. Regarding method coverage, specific methods like AHP, ARAS, ELECTRE, and SAW enjoy widespread support. In contrast, more specialized or less commonly used methods like MARE, SECA, and SRMP are exclusive to specific libraries. This hints at their specialized nature or limited application in the MCDA field. Implication-wise, pyDecision, and JMcDM are all-encompassing choices for generalists, covering various classic and fuzzy methods. Specialists might find libraries like R MCDA, pyFDM, and pyIFDM more aligned with their needs, focusing on specific subsets of MCDA methods. However, it's crucial to acknowledge the limitations inherent in existing MCDA software. The predilection for single-model approaches can limit versatility, adapting to a broad spectrum of decision-making scenarios can be challenging, and user-friendliness remains an obstacle, particularly for MCDA novices. Therefore, while each library brings its unique advantages to the table, the choice of the most appropriate tool should be carefully considered based on the specific needs and constraints of the decision-making environment.

## 3. PyDecision Methods and Capabilities - First Case Study

In this section, we aim to demonstrate the versatility of pyDecision in conjunction with ChatGPT. The first case study explores various methods capable of generating rankings, including two methods from the academic literature not natively supported by pyDecision. Then, ChatGPT plays an integral role in interpreting the outcomes. We employ a sequence of targeted prompts to guide the decision-maker in evaluating and understanding the results. One can note that the underlying analysis would essentially be the same since fuzzy methods produce crisp rankings even with fuzzy input. Therefore, we will not conduct a separate evaluation for ranks generated by fuzzy methods to avoid redundancy.

The first Case Study can be accessed at < https://bit.ly/47aAVrZ >

### 3.1 First Case Study – Methods Comparison

The first case study involves selecting an optimal material for a cryogenic storage tank destined to transport liquid nitrogen. The specific performance requirements of the storage tank were meticulously translated into corresponding material attributes. The chosen material must demonstrate excellent weldability, processability, lower density, and specific heat metrics, among the essential qualities. It must also present a minimal thermal expansion coefficient and thermal conductivity while maintaining sufficient strength, stiffness, and toughness at the operating temperature. For this particular application, materials showcasing higher mechanical properties are sought after, with emphasis on the highest values in toughness index (TI), yield strength (YS), and Young's modulus (YM). The selection process evaluates seven distinct attributes, representing the desired material properties against seven alternative materials. Among these, TI, YS, and YM are beneficial attributes (max), where higher values are advantageous. Conversely, the remaining four attributes - namely, density (D), thermal expansion (TE), thermal conductivity (TC), and specific heat (SH) - are deemed non-beneficial (min), as their lower values align more closely with the application's needs (RAO, 2006).

The alternatives considered for the material selection, as outlined in Table 2, include a1 (Material 1 - Al 2024-T6), a2 (Material 2 - Al 5052-O), a3 (Material 3 - SS 301-FH), a4 (Material 4 - SS 310-3AH), a5 (Material 5 - Ti-6Al-4V), a6 (Material 6 - Inconel 718), and a7 (Material 7 - 70Cu-30Zn). To tailor the selection process to the application's specific needs, Table 2 also shows the weights Rao (2006) and Manshadi et al. (2007) allocated to reflect each attribute's relative importance. These weights function as pivotal parameters in assessing each material's suitability, steering the decision-making toward alignment with the application's distinct requirements.

The dataset, as displayed in Table 2, reveals the complexity of the decision-making process, as no single material excels across all the evaluated attributes. This intricate balance of competing properties underscores the challenge of material selection for this highly specialized task.

**Table 2 – First Case Study**

| | TI (MAX) | YS (MAX) | YM (MAX) | D (MIN) | TE (MIN) | TC (MIN) | SH (MAX) |
|---|---|---|---|---|---|---|---|
| **weigths** | 0.28 | 0.14 | 0.05 | 0.24 | 0.19 | 0.05 | 0.05 |
| **a1** | 75.50 | 420 | 74.2 | 2.80 | 21.40 | 0.370 | 0.16 |
| **a2** | 95 | 91 | 70 | 2.68 | 22.10 | 0.330 | 0.16 |
| **a3** | 770 | 1365 | 189 | 7.90 | 16.90 | 0.040 | 0.08 |
| **a4** | 187 | 1120 | 210 | 7.90 | 14.40 | 0.030 | 0.08 |
| **a5** | 179 | 875 | 112 | 4.43 | 9.40 | 0.016 | 0.09 |
| **a6** | 239 | 1190 | 217 | 8.51 | 11.50 | 0.310 | 0.07 |
| **a7** | 273 | 200 | 112 | 8.53 | 19.90 | 0.290 | 0.06 |

Table 3 presents the rankings of each alternative as determined by two distinct techniques: the Graph Theory technique, as proposed by Rao (2006), and a Non-linear Normalization method enhanced by a modified digital logic approach, as described by Manshadi et al. (2007).

**Table 3 – First Case Study Results**

| | Rao (2006) | Manshadi et al. (2007) |
|---|---|---|
| **a1** | 17.29 (6th) | -1.71 (5th) |
| **a2** | 16.26 (7th) | -8.75 (7th) |
| **a3** | 39.11 (1st) | 47.40 (1st) |
| **a4** | 30.63 (3rd) | 31.88 (4th) |
| **a5** | 34.05 (2nd) | 43.52 (2nd) |
| **a6** | 29.04 (4th) | 33.44 (3rd) |
| **a7** | 20.04 (5th) | -3.07 (6th) |

Regarding the ranks for each alternative, Table 3 shows appealing convergences and divergences. For example, both methods unanimously ranked the third alternative as the best (1st place); the same goes for the second and fifth alternatives. However, there are variations in the middle rankings; the first alternative was ranked 6th according to Rao (2006) but fared slightly better at 5th place in Manshadi et al. (2007).

Table 4 is structured to display a comprehensive comparison of various MCDA methods. The rows list the methods under consideration. The columns represent the different alternatives being evaluated. Within the cells, the rank is assigned for each alternative. The following methods were used: ARAS, CoCoSo, CODAS COPRAS, CRADIS, EDAS, GRA, MABAC, MACBETH, MAIRCA, MARCOS, MAUT, MOORA, MOOSRA, MULTIMOORA, OCRA, ORESTE, PIV, PROMETHEE II, PROMETHEE IV, EC PROMETHEE, PSI, ROV, SAW, SPOTIS, TODIM, TOPSIS, VIKOR, WSM, WPM, and WASPAS. Additionally, the results obtained by Rao (2006) and Manshadi et al. (2007) were included.

Specific parameters are relevant to particular methods, as follows: For PROMETHEE II, PROMETHEE IV, or EC PROMETHEE: Q = 5, 10, 1.7, 0.02, 0.01, 0.01, 0.01; S = 7, 15, 1.9, 0.03, 0.02, 0.02, 0.02; P = 9, 20, 2.5, 0.04, 0.03, 0.03, 0.03; F= Usual Preference. For EC PROMETHEE additionally: Custom Sets = 0.5, 0.5, 0.5, 0.5, 0.5, 0.5, 0.5; Iterations = 10000. For MAUT: Utility Functions = Exponential / Step Functions (Max / Min criterion); Step Size = 1; For CoCoSo: L = 0.5. For CODAS: Lambda = 0.02. For GRA: Epsilon = 0.5 For ORESTE: Alpha = 0.4. For TODIM: Teta = 1. For VIKOR: Strategy Coefficient = 1. For SPOTIS: Smin = 70, 90, 50, 2.7, 9.0, 0.01, 0.05; Smax = 780, 1200, 220, 9.7, 24.0, 0.41, 0.25.

**Table 4 – Rank Comparison**

| | a1 | a2 | a3 | a4 | a5 | a6 | a7 | | a1 | a2 | a3 | a4 | a5 | a6 | a7 |
|---|---|---|---|---|---|---|---|---|---|---|---|---|---|---|---|
| Rao (2006) | 6 | 7 | 1 | 3 | 2 | 4 | 5 | OCRA | 6 | 7 | 1 | 2 | 3 | 4 | 5 |
| Manshadi et al. (2007) | 5 | 7 | 1 | 4 | 2 | 3 | 6 | ORESTE | 5 | 6 | 2 | 3 | 1 | 4 | 7 |
| ARAS | 5 | 6 | 1 | 4 | 2 | 3 | 7 | PIV | 5 | 6 | 1 | 4 | 2 | 3 | 7 |
| CoCoSo | 5 | 6 | 1 | 3 | 2 | 4 | 7 | PROMETHEE II | 7 | 6 | 1 | 4 | 2 | 3 | 5 |
| CODAS | 4 | 2 | 1 | 6 | 3 | 5 | 7 | PROMETHEE IV | 6 | 7 | 1 | 3 | 4 | 2 | 5 |
| COPRAS | 5 | 7 | 1 | 4 | 2 | 3 | 6 | EC PROMETHEE | 5 | 7 | 1 | 3 | 2 | 4 | 6 |
| CRADIS | 6 | 7 | 2 | 3 | 1 | 4 | 5 | PSI | 5 | 6 | 2 | 4 | 1 | 3 | 7 |
| EDAS | 6 | 7 | 1 | 4 | 2 | 3 | 5 | ROV | 5 | 6 | 1 | 4 | 2 | 3 | 7 |
| GRA | 5 | 4 | 1 | 6 | 2 | 3 | 7 | SAW | 5 | 6 | 1 | 4 | 2 | 3 | 7 |
| MABAC | 5 | 6 | 1 | 3 | 2 | 4 | 7 | SPOTIS | 5 | 6 | 1 | 4 | 2 | 3 | 7 |
| MACBETH | 5 | 6 | 1 | 4 | 2 | 3 | 7 | TODIM | 5 | 6 | 1 | 3 | 2 | 4 | 7 |
| MAIRCA | 5 | 6 | 1 | 4 | 2 | 3 | 7 | TOPSIS | 5 | 6 | 1 | 4 | 2 | 3 | 7 |
| MARCOS | 5 | 6 | 1 | 4 | 2 | 3 | 7 | VIKOR | 3 | 5 | 6 | 4 | 1 | 2 | 7 |
| MAUT | 4 | 6 | 1 | 3 | 5 | 2 | 7 | WSM | 5 | 6 | 1 | 4 | 2 | 3 | 7 |
| MOORA | 5 | 6 | 1 | 4 | 2 | 3 | 7 | WPM | 5 | 6 | 1 | 3 | 2 | 4 | 7 |
| MOOSRA | 5 | 7 | 1 | 4 | 2 | 3 | 6 | WASPAS | 5 | 6 | 1 | 3 | 2 | 4 | 7 |
| MULTIMOORA | 5 | 7 | 1 | 3 | 2 | 4 | 6 | -//- | - | - | - | - | - | - | - |

Upon analyzing Table 4, several patterns and variations can be observed. The alternative a3 has the most stable rank across almost all methods. It is frequently ranked at position 1, indicating a strong consensus that it's the best alternative according to the evaluated criteria. Alternatives a2 and a7 show a moderate variation in their ranks; while they are generally ranked towards the middle or bottom across methods, the exact position fluctuates. For instance, for a2, rankings range from 2 (in CODAS) to 7 (in several methods). Conversely, in the alternatives a4 and a5, the rank varies extensively among the methods. For example, a4 is ranked anywhere from 3rd to 6th place, and a5 moves between the 1st and 4th positions, indicating less consensus on assessing these alternatives.

Should the decision-maker desire, they can derive a final rank using the predominant ranks for each alternative. However, caution is advised in multimodal scenarios, where an alternative may possess multiple ranks, or in tie situations where different alternatives share identical ranks. In our case study, the final rank obtained by the mode is a3, a5, a6, a4, a1, a2, and a7. Alternatively, a voting system can be employed to ascertain ranks. Within pyDecision, Borda and Copeland's methods are natively available - for this particular case study - both methods have computed the same final rank as the mode. Nevertheless, the decision-maker retains the flexibility to employ any other preferred method.

### 3.2 First Case Study – Methods Comparison – Large Language Model Aid

To extract deeper insights and potentially uncover hidden patterns in Table 4, we'll leverage the analytical capabilities with ChatGPT (model gpt4). We plan to use a sequence of targeted prompts to augment our initial analysis by shedding light on aspects we might have overlooked.

**Prompt 1**: *Which methods are more similar and which ones are more dissimilar?*

**Response**: *The similarity of methods can be analysed by looking at how consistently they rank the alternatives. From the given outranking table, it is not possible to declare precisely which methods are more similar and which are more dissimilar without additional calculations (e.g., calculating Euclidean distance or Pearson's correlation). However, some general trends can be observed. For instance, within the methods, "aras" and "macbeth," the provided ranking values in the table appear to show a high degree of similarity across all alternatives as they both rank alternatives the same way. In the same way, "mabac" and "marcos" or "maut" and "moora" methods seem to give similar rankings across the alternatives. On the other hand, a method like "vikor" appears to be more dissimilar from others. For example, in "vikor," the first and fifth alternatives are given a rank of 6th and 1st, which differs significantly from most other methods. Nevertheless, this is a preliminary assessment, and a more accurate interpretation would necessitate mathematical computations such as calculating correlation coefficients to determine the extent* *of similarity or dissimilarity. This also assumes that the ordering of the alternatives is all that matters, and not the actual values attached to each position.* [sic]

**Prompt 2**: *Which alternative(s) consistently ranks high or low across all the methods??*

**Response**: *The alternative that consistently ranks high across all the methods is the third one. It ranks first in almost all methods except two (rov and oreste), where it ranks second. On the other hand, the alternative that consistently ranks low across all the methods is the sixth one. It mostly ranks seventh, which is the lowest, with a few exceptions where it ranks slightly higher.* [sic]

**Prompt 3**: *Are there any noticeable differences in rankings across the methods? if so, what could account for these differences??*

**Response**: *Yes, there are noticeable differences in rankings across the multiple criteria decision analysis (MCDA) methods. The third alternative, for instance, consistently ranks first in almost all MCDA methods - except for a few like OCRA, PIV, and SAW where it ranks second or lasts sixth in VIKOR - indicating a strong performance across multiple evaluation standards. The sixth alternative, on the other hand, mostly ranks in the middle. Still, there are also few methods like MOORA, TODIM, and COPRAS where it achieves weaker positions, and a better position in EC_Promethee & WASPAS. The same goes for the other alternatives, where one can notice variations in their positions depending on the applied MCDA method. These discrepancies may be due to different prioritization and analysis mechanisms intrinsic to each MCDA method. Depending on the method, various aspects such as weight assignment, type of optimization (Max or Min), preference analysis, consideration of subjective opinions, normalization methods, etc., could have different impacts on the ranking. For instance, some methods may assign more weight to certain criteria, or directly incorporate the decision-maker's preferences into the decision-making process, thus influencing the final outcomes. Others could use an additive aggregation function, while others could employ a multiplicative aggregation function, leading to alternate outcomes. In other words, the specific methodology of each MCDA technique and their differing treatment of both the criteria and the alternatives' performances on these criteria could result in different rankings. In addition, uncertainty in original data, potential biases in data collection and treatment tools, and subjectivity in interpreting the criteria could also contribute to differences in the final ranking. Moreover, method-related factors, such as sensitivity to changes in parameters, handling of inequality among the criteria, and the ability to consider interactions among criteria - along with problem-related aspects, such as complexity of the problem, number and nature of alternatives and criteria, and conflict levels among criteria, could also play a role.* [sic]

**Prompt 4**: *Is there a consensus among the methods for any specific alternative(s)?*

**Response**: *If we calculate the mean (average) ranking for each alternative across all MCDA methods, we get a consensus that can be used as a comprehensive ranking for the alternatives. Based on the given outranking table, the average ranks are: - Alternative 1: Mean Rank = 5.24 - Alternative 2: Mean Rank = 6.16 - Alternative 3: Mean Rank = 1.26 - Alternative 4: Mean Rank = 3.84 - Alternative 5: Mean Rank = 2.03 - Alternative 6: Mean Rank = 3.57 - Alternative 7: Mean Rank = 6.39 Therefore, the consensus among the methods could be ordered as follows: 1. Third Alternative 2. Fifth Alternative 3. Sixth Alternative 4. Fourth Alternative 5. First Alternative 6. Second Alternative 7. Seventh Alternative This particular ordering means alternative 3 is the most preferred, alternative 5 is the next most preferred, and so on. The seventh alternative is the least preferred based on the ranking by these MCDA methods.* [sic]

**Prompt 5**: *Are there any unexpected rankings for certain alternatives when comparing across methods??*

**Response**: *Yes, there are some unexpected rankings when comparing across different MCDA methods. Here are few instances: 1. In the Second alternative, there is a variation in the ranking of CODAS and MARCOS which placed it at 2nd position whereas all other methods placed it at 6th or 7th position. 2. In the Third alternative, there is a significant deviation in VIKOR's ranking, which placed it at 6th position, whereas all other methods placed it at 1$^{st}$ position. 3. Similarly in the Fourth alternative, the ranking by CODAS and*

*GRA method is deviating from other methods. 4. In the Fifth alternative, MAUT has ranked the alternative as 5th, while most other methods rank it as 2nd. 5. The Seventh alternative shows great discrepancy. It was ranked 7th by most methods, but ranked 5th by some like GRA and Promethee IV. Such discrepancies and unexpected rankings could be because each MCDA method uses a different set of criteria weights or preference function, and they might approach the problem from different perspectives or assumptions.* [sic]

**Prompt 6**: *Are there any methods that consistently rank alternatives differently than most other methods??*

**Response**: *From the given outranking table, we can see that some Multicriteria Decision Making Aid (MCDA) methods consistently rank alternatives differently than most other methods. For instance: - 'codas' consistently ranks the second alternative as the best (2nd rank), while most other methods rank it anywhere between 6th to 7th. - 'cradis' method, alternately, ranks the third alternative as second-best (2nd rank), whereas the majority of other methods rank it as the best (1st rank). - 'vikor' method ranks the first alternative as 3rd and the third alternative as 6th, which is quite distinct from other methods that rank those alternatives as 6th and 1st respectively. It is also important to mention, not only these methods predict different ranking but also the degree of differences in their results vary. Some methods rank certain options significantly differently than other methods, while some methods show relatively smaller differences. The degree of difference is important information for decision-makers, because it reflects the level of uncertainty or disagreement about the ranking of the alternatives. It might be beneficial to use multiple methods for decision-making in order to gain comprehensive insights and to consider various perspectives. Additionally, it's recommended to understand the underlying principles and assumptions of each MCDA method as they significantly influence the ranking output. Different methods may yield different results due to these operating principles and assumptions, offering different insights into the decision problem.* [sic]

**Prompt 7**: *What is the most common ranking for each alternative across all methods??*

**Response**: B*ased on the table given, below is the most common ranking for each alternative: 1. The most common ranking for the first alternative is 5. 2. The most common ranking for the second alternative is 6. 3. The most common ranking for the third alternative is 1. 4. The most common ranking for the fourth alternative is 3 or 4. 5. The most common ranking for the fifth alternative is 2. 6. The most common ranking for the sixth alternative is 3 or 4. 7. The most common ranking for the seventh alternative is 7.* [sic]

Utilizing ChatGPT to analyze the data from Table 4 has added a nuanced layer of understanding that extends beyond what traditional analytical methods offer. The model provided insights into the congruencies and divergences across various MCDA methods, showcasing its value for initial data interpretation and hypothesis formation. Responses to specific prompts like Prompt 4, which discussed a consensus among methods based on mean ranks, require caution since averaging ordinal rankings can be misleading. Prompt 7 shows the most commonly occurring ranks (mode) and is the correct approach to dealing with ordinal data. It's crucial to remember that while ChatGPT provides beneficial initial insights, it doesn't replace the need for rigorous analytical scrutiny. Researchers should use the model as a complement to, not a replacement for, conventional research methods.

### 3.3 First Case Study – Correlation Analysis

To examine the relationships between different methods more deeply, as presented in Table 4, we will employ the Kendall Tau correlation coefficient for our analysis. This choice is particularly apt given that Kendall Tau is primarily designed for ordinal data, aligning well with the nature of this problem. Figure 1 offers a color-coded matrix to visualize these correlations. Positive relationships between methods are indicated by varying shades of blue, with darker blue signifying stronger positive correlations. Conversely, shades of green and yellow represent correlations near zero, highlighting the less positive relationships between methods. Notably, the matrix reveals no negative correlations between the methods, underscoring a general agreement trend, albeit to varying degrees.

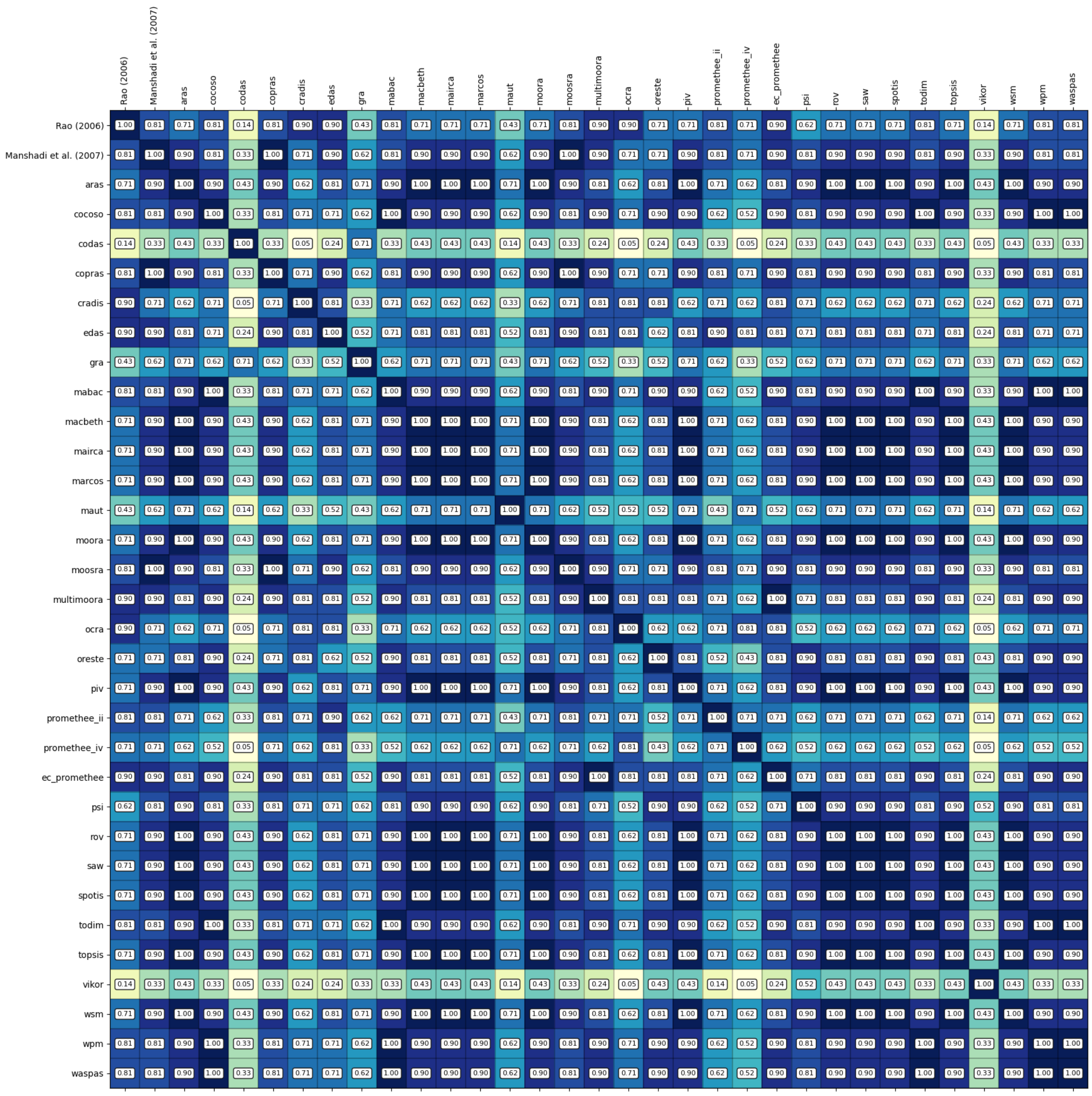

**Figure 1 – First Case Study Rank Correlation**

Figure 1 provides a comprehensive view of the correlations between the various MCDA methods. Excluding the diagonal values, which naturally stand at 1.0, the overall minimum correlation observed was 0.05, while the maximum reached 1.0. Notably, the methods CODAS and VIKOR display the least congruence with other methods and between themselves, both holding a correlation of 0.05, indicating that these methods might offer unique perspectives compared to the others in the matrix.

## 3.4 First Case Study – Correlation Analysis – Large Language Model Aid

To extract deeper insights and potentially uncover hidden patterns in Figure 1, we'll leverage the analytical capabilities with ChatGPT (model gpt4). We plan to use a sequence of targeted prompts to augment our initial analysis by shedding light on aspects we might have overlooked.

**Prompt 1**: *Explain the significance of analyzing the correlation of ranks between different MCDA methods.*

**Response**: *The purpose of analyzing the correlation of ranks between different MCDA (Multicriteria Decision Analysis) methods is to understand how consistent the methods are with each other. In other words, it establishes the degree to which one method tends to give the same rankings as another method. Looking at the correlation table: - High correlation values (near 1) between any two methods mean that the two methods rank alternatives similarly. For instance, 'Rao (2006)' and 'Manshadi et al. (2007)' show a correlation value of 0.809524 indicating a strong positive agreement between these two. - Low correlation values (near 0) indicate that the two methods often rank alternatives very differently. For instance, 'Rao (2006)' and 'Codas' show a correlation value of 0.142857 indicating low agreement between these two. This analysis helps researchers or decision-makers understand which methods are most consistent with each other, which could be important when choosing which method to use or when interpreting results. If certain methods consistently give similar rankings, a researcher might feel confident using any of them. However, if some methods often disagree with others, it could be an indication that these methods are taking into account some factors that others are not, or weighing some factors more heavily. In this context, understanding these inconsistencies and the reasons for them could yield important insights about the limitations and advantages of different MCDA methodologies. Moreover, if the decision is sensitive or critical, decision-makers may opt to use the most consistent methods or a set of methods that produce similar rankings to ensure the robustness and reliability of the chosen solution.* [sic]

**Prompt 2**: *What are the implications if there is a high correlation between the ranks produced by different MCDA methods?*

**Response**: *A high correlation between ranks produced by different Multicriteria Decision Analysis (MCDA) methods would suggest a level of agreement or consistency in the decision-making process. This can have several implications: 1. Reveals Robustness: A high correlation among different methods signifies robust results. This means the decision or rank obtained is not sensitive to the specific MCDA method used. 2. Confirms Validity: If multiple methods are showing similar outcomes, it adds to the validity of the results. This can enhance trust and confidence in the outcomes of the decision-making process. 3. Simplifies Process: Knowing that different methods yield very similar results can simplify the decision-making process. Decision-makers might choose to utilize one method over another based on ease, simplicity, or computational efficiency without significantly impacting the result. 4. Indicates Redundancy: However, in case you employed various MCDA methods to ensure an extensive and diverse analysis of decision-making factors, high correlation could mean redundancy. If multiple methods provide very similar results, it may not be necessary to implement all of them. 5. Agreement in outcomes: A high correlation indicates good agreement between the rankings. Any selected action or trend will probably be the same or very similar no matter which MCDA method is used. Finally, high correlation between ranks doesn't necessarily guarantee high correlation between the actual numerical scores given, as the methods for arriving at these can be very different depending on the MCDA method used.* [sic]

**Prompt 3**: *What might cause a low correlation in rankings between different MCDA methods?*

**Response**: *A low correlation between different MCDA methods might result from several factors: 1. Different MCDA methods may take into account different aspects of the alternatives, hence reflecting divergent perspectives in the decision-making process. 2. The methods analyzed might implement completely different theoretical models, hence leading to dissimilar conclusions. 3. The divergence can also occur due to the different assumptions that each method makes about the decision problem. 4. Additionally, the chosen outputs of the methods might vary greatly. Some methods may output complete rankings, while others might classify decision alternatives into preference groups. 5. The methods may also weight the criteria differently, and small changes in weights can substantially modify the final ranking. Some methods are more sensitive to these weights than others. 6. Lastly, the methods might define and interpret uncertainty* *differently. This can make a huge difference in certain scenarios where the data contains significant uncertainties.* [sic]

**Prompt 4**: *What precautions or considerations should be taken when comparing the rankings of different MCDA methods?*

**Response**: *When comparing the rankings of different Multicriteria Decision Analysis (MCDA) methods, it is important to bear the following considerations in mind: 1. Not all MCDA methods evaluate criteria in the same way. Some methods may give total priority to the best performing alternative, while others play with compensations. 2. An MCDA method may not work in the same way for all kinds of problems. For instance, some MCDAs require a linear preference function, meaning that they are not suited to non-linear problems. 3. Different MCDA methods may involve different assumptions and logic. For instance, some may assume compensation between criteria, whereas others do not, etc. 4. The results of each MCDA method can markedly vary depending on how the analysis was set up. For instance, the weight assigned to different criteria can have a major impact on results. 5. It's important to take into account that correlation values are not precise indicators of similarity between rankings produced by different methods. Given such considerations, rather than comparing the end ranking of different MCDA methods directly, it may be more informative to examine the process that each method uses to arrive at its ranking. This may give you a better appreciation of why different methods produced different results and more confidence in the method that best matches your intuition and understanding of the problem at hand.* [sic]

Utilizing ChatGPT with a researcher's expertise can offer fresh perspectives on a given issue. However, it's vital to recognize its limitations in capturing every nuance of a broad topic. Specific and directed prompts tend to yield more targeted responses. In contrast, open-ended inquiries might miss certain critical aspects, as observed in the case of the discrepancies between VIKOR and CODAS compared to other methods. Despite the insights provided by ChatGPT, human oversight remains indispensable, especially when discerning evident patterns like those in our example.

## 4. PyDecision Methods and Capabilities – Second Case Study

Again, this section aims to demonstrate the versatility of pyDecision in conjunction with ChatGPT. The second case study explores various methods capable of generating weights, including two methods from the academic literature not natively supported by pyDecision. Then, ChatGPT plays an integral role in interpreting the outcomes. We employ a sequence of targeted prompts to guide the decision-maker in evaluating and understanding the results. In this section, we showcase the combination of pyDecision and ChatGPT. Our second case study examines diverse weight-generating methods, also featuring two techniques from academic literature not inherently integrated into pyDecision. Again, ChatGPT is used through a series of prompts, and we try to comprehend the results better.

The second Case Study can be accessed at < https://bit.ly/3QAaJle >

### 4.1 Second Case Study – Methods Comparison

The second case study is based on the research conducted by Bottero et al. (2015), where they developed a framework for multicriteria decision-making utilizing the ELECTRE III method. The study employed this framework based on six criteria to assess five projects labeled a1 to a5 thoroughly. Investment Costs (IVC) is the first criterion focusing on construction expenditures. The criterion of Profitability (PRF) follows, which evaluates not only the direct financial gains but also the broader economic benefits of a project, using a qualitative scale. Likewise, New Services for the Population (NSP) measures a project's contributions to communal amenities such as recreational spaces or green areas. Landscape Ecology (LSE) examines a project's ecological footprint, precisely its effect on local biodiversity, quantified in hectares of naturalized land. Environmental Effects (EVE) scrutinizes the potential impacts on physical and ecological aspects, like hydrology and geotechnical conditions, also employing a qualitative scale. Lastly, the binary criterion of Consistency with Local Planning Requirements (CPR) ascertains if a project adheres to existing regulatory frameworks. All the criteria are designed for maximization except for Investment Costs (IVC), which is a minimization criterion.

**Table 5 – Second Case Study**

| | IVCI (MIN) | PRF (MAX) | NSP (MAX) | LSE (MAX) | EVE (MAX) | CPR (MAX) |
|---|---|---|---|---|---|---|
| **a1** | 30,000 | 3 | 1 | 2 | 4 | 1 |
| **a2** | 45,000 | 3 | 5 | 5 | 5 | 1 |
| **a3** | 90,000 | 1 | 6 | 3.2 | 7 | 1 |
| **a4** | 120,000 | 1 | 7 | 3.5 | 6 | 1 |
| **a5** | 900,000 | 7 | 7 | 1 | 3 | 0 |

Table 5 shows the values that represent the performance of each project (a1-a5) against these criteria. For instance, project a1 has an investment cost of 30,000 and scores 3 in profitability, while project a5 has an investment cost of 900,000 but ranks highest in environmental effects with a score of 7. Table 6 delineates the weights assigned to the criteria based on two distinct methods: Bottero et al. (2015) and Rodrigues et al. (2021). For each criterion, weights signify its importance in the overall decision-making process. Rodrigues et al. (2021) employed an algorithm named Ranking Trees to calculate the weights for each criterion. In contrast, Bottero et al. (2015) relied on domain specialists to determine the weightage.

**Table 6 – Second Case Study - Weights**

| | Bottero et al. (2015) | Rodrigues et al. (2021) |
|---|---|---|
| **IVCI** | 0.140 | 0.170 |
| **PRF** | 0.250 | 0.130 |
| **NSP** | 0.080 | 0.100 |
| **LSE** | 0.200 | 0.230 |
| **EVE** | 0.310 | 0.190 |
| **CPR** | 0.020 | 0.180 |

Table 7 shows the MCDA methods BWM, CILOS, CRITIC, Entropy, IDOCRIW, and MEREC; additionally, two custom methods have been integrated, Bottero et al. (2015) and Rodrigues et al. (2021). For BWM, there's an emphasis on identifying the most important (MIC = 2, 4, 5, 3, 1, 6) and least important (LIC = 6, 1, 3, 5, 4, 2) criteria, represented as rankings from 1 to 6 for each criterion in each situation; therefore the best criterion is EVE (1$^{st}$ in MIC) and the worst criteria PRF(1$^{st}$ in LIC).

**Table 7 – Second Case Study - Results**

| | IVCI | PRF | NSP | LSE | EVE | CPR |
|---|---|---|---|---|---|---|
| Bottero et al. (2015) | 0.140 | 0.250 | 0.080 | 0.200 | 0.310 | 0.020 |
| Rodrigues et al. (2021) | 0.170 | 0.130 | 0.100 | 0.230 | 0.190 | 0.180 |
| BWM | 0.178 | 0.119 | 0.084 | 0.124 | 0.416 | 0.078 |
| CILOS | 0.050 | 0.059 | 0.361 | 0.08 | 0.116 | 0.327 |
| CRITIC | 0.140 | 0.313 | 0.178 | 0.106 | 0.114 | 0.148 |
| Entropy | 0.288 | 0.236 | 0.116 | 0.109 | 0.039 | 0.212 |
| IDOCRIW | 0.361 | 0.166 | 0.117 | 0.114 | 0.086 | 0.156 |
| MEREC | 0.065 | 0.055 | 0.074 | 0.028 | 0.014 | 0.763 |

The second case study shows that BWM has a unique weight distribution, emphasizing the EVE criterion. CILOS seems to emphasize NSP and CPR more, while CRITIC prioritizes PRF. Entropy allocates the highest weight to IVCI, while IDOCRIW does the same but with an even more pronounced emphasis. Lastly, MEREC stands out by assigning most of its weight to CPR.

## 4.2 Second Case Study – Methods Comparison – Large Language Model Aid

To extract deeper insights and potentially uncover hidden patterns in Table 7, we'll leverage the analytical capabilities with ChatGPT (model gpt4). We plan to use a sequence of targeted prompts to augment our initial analysis by shedding light on aspects we might have overlooked.

**Prompt 1**: *Which methods are more similar and which ones are more dissimilar?*

**Response**: *To determine the similarity and dissimilarity between different MCDA methods, various statistical analysis methods can be used. Among those, correlation analysis and clustering analysis are the most commonly used. For this specific type of data (weight of each criterion calculated by each MCDA method), correlation analysis would be a good choice. This analysis measures the statistical relationship, but not deterministic, between two sets of variables. High positive correlation value (close to 1) means two methods are very similar - they tend to give higher weights to the same criteria and lower weights to the same set of other criteria. On the other hand, high negative correlation value (close to -1) means two methods are very dissimilar - they tend to give a higher weight to a criterion that the other method gives lower weight, and vice versa. However, from the table alone, we can't definitively state which methods are most similar or dissimilar as it would require computation and analysis. As per the rough estimate based on the provided weights, it looks like 'Bottero et al. (2015)' and 'Rodrigues et al. (2021)' are somewhat similar. The 'cilos' method seems to be relatively dissimilar comparing with others, based on its distinct patterns of weights assignment. But please note that these are only heuristic observations. For an accurate similarity/dissimilarity evaluation, quantifiable statistical measures such as correlation coefficients should be computed.* [sic]

**Prompt 2**: *Are there certain criteria that consistently receive high weights across all methods? What might these key criteria suggest about the decision problem at hand?*

**Response**: *The information from the table seems to be a bit incomplete for providing in-depth explanations. However, based on what's given, it seems that there might be certain criteria that receive high weights across different methods when averaged. For example, first & fifth criteria average to relatively high weights across all methods, suggesting they might be particularly important for the decision problem at hand. In MCDA (multiple-criteria decision analysis), the weights assigned to different criteria reflect their relative importance in the decision problem. Hence, these weights suggest that these two factors/criteria may be of considerable significance, potentially impacting the overall decision to a great extent. However, the specifics about what these criteria represent or the decision problem at hand are not given in the question, which limits the analysis depth. That said, keep in mind that even though a criterion might receive high weights on average, it doesn't mean it's always the most significant– different methods may prioritize different aspects, reflecting varying perspectives on the same decision problem. Additionally, the decision-making context (whether you're selecting the best alternative, ranking alternatives, sorting them into categories, etc.) also affects the interpretation of such weights. The comparison of the weights across different methods might be used to validate the robustness of the results (if similar weights are obtained, confidence in these can be increased) or to explore how sensitive the results are to the specific weighting method used. Note that the high weight in the last column (0.763368) is potentially an outlier or it might be indicating a very significant criterion for merec method. Without contextual information, it is impossible to provide definite explanations.* [sic]

**Prompt 3**: *Conversely, are there criteria that consistently receive low weights across all methods? This could indicate aspects that are less important to the decision context.*

**Response**: *It appears that the later methods (cilos, critic, entropy, idocriw, merec) consistently attribute lower weights to the 5th criterion (row 5), with merec attributing the lowest weight of all (0.013805). This suggests that the 5th criterion may be of less importance in these MCDA methods. Interestingly, the earlier methods (Bottero et al. 2015, Rodrigues et al. 2021, bwm) attribute more importance to the 5th criterion. This could indicate a difference in prioritization between different methods. Also, the 6th criterion seems to receive a very low weight from the methods Bottero et al. (2015) and Rodrigues et al. (2021), which might mean that these methods don't consider the 6th criterion to be very important. However, further analysis* *would be needed to determine whether these patterns hold true across different decision contexts and to interpret what this means in each specific context.* [sic]

**Prompt 4**: *How much variability is there in weights assigned to each criterion by different methods? High variability could suggest that different methods interpret the importance of the criteria differently.*

**Response**: *To determine the variability in weights assigned to each criterion by different methods, it would be useful to calculate the standard deviation for each row. The standard deviation will show how much variation exists from the average (mean) weight. For example, let's calculate the standard deviation of the weights assigned to the first criterion by each method: - Value by Bottero et al. (2015): 0.14 - Value by Rodrigues et al. (2021): 0.17 - Value by bwm: 0.178310 - Value by cilos: 0.039830 - Value by critic: 0.140939 - Value by entropy: 0.288163 - Value by idocriw: 0.360795 – Value by merec: 0.065421 We could use a standard deviation calculator or a software like Excel for this calculation. Then, we'd repeat this process for each row (i.e., each criterion) to see how much variability exists in the importance assigned to each criterion by the different methods. After obtaining all the standard deviations, if they're high, then it suggests that the methods interpret the importance of the criteria differently. But if they're low, then it suggests that the methods interpret the importance of the criteria similarly. Note that I did not provide the actual standard deviation calculations in these examples because this assistant doesn't have access to computational functions for calculating this.* [sic]

**Prompt 5**: *Are there any noticeable correlations between the weights assigned by different methods*

**Response**: *Based on the data provided, it's hard to determine any solid correlations just by visual assessment. However, few patterns can be observed: 1. Bottero et al. (2015) and Rodrigues et al. (2021): The weightings for the third, forth and fifth criteria from these two methods tend to move in the same direction, i.e., when one increases the other increases as well, and when one decreases the other also decreases. 2. Bottero et al. (2015) and Cilos: For the first and third criterias, when Bottero et al. (2015) has a low rating, Cilos displays a high rating, suggesting an inverse relationship. 3. Rodrigues et al. (2021) and IDOCRIW: For the first and second criterias, when Rodrigues et al. (2021) moves in one direction, IDOCRIW moves in opposite directions. To conclude on any correlations and their significance, a statistical analysis (Pearson correlation, Spearman correlation, etc.) would be needed to analyze the numerical patterns within and between the different MCDA methods. This includes calculating correlation coefficients which could lend a quantitative understanding to any relationships between these methods. AI models would not be able to conclusively identify correlations through purely visual inspection of the data.* [sic]

**Prompt 6**: *Can you identify outlier methods that assign weights significantly different from others?*

**Response**: *To identify the outlier MCDA methods, one way is to compare each method's weights to the median (or mean) of all the methods for each criterion. Let's consider outlier methods as those which deviate from the others by more than a certain threshold, say 0.2. Bottero et al. (2015) and Rodrigues et al. (2021) seem to assign similar weights to each criterion. Most of their weights fall in the same range. For 'bwm', weights are more variable. The weight of the fifth criterion is the highest among all the methods, which could make 'bwm' an outlier for this specific criterion. 'cilos' seems to assign significantly different weights to the third and sixth criteria, which are way higher than the other methods. And so, it can be considered an outlier for these criteria. 'critic' falls in the same pattern as 'bwm' with higher weights for the second criterion. 'entropy' assigns the most weight to the first criterion which is higher than other methods. 'idocriw' has relatively low variability in weights across the criteria but still the first criterion has the highest weight which is a bit higher than others. 'merec' features a very high weight for the last criterion, which could make it an outlier method for this criterion. It's important to note that the relevance of these differences depends on the domain of application, and on the specific method's inherent characteristics. There may be good reasons for these differences, and they may or may not indicate actual problems.* [sic]

**Prompt 7**: *Is there a specific method that consistently assigns higher or lower weights to all criteria? If so, what does this indicate about the method's evaluation approach?*

**Response**: *Based on the table given, it's hard to pin down a method that consistently assigns higher or lower weights to all criteria without knowing the specific criteria being evaluated. However, conceptually, if a method assigns consistently higher weights to all criteria, it may handsomely value all aspects being measured, making it a broad-brush approach that might be good for general analyses but lack specificity. In contrast, if a method consistently assigns lower weights to all criteria, it may indicate that the method is more conservative or restrictive in its weighting, with a more skeptical approach in acknowledging the impact of each criterion. Observations from the table do indicate that, 'idocriw' can be seen to assign relatively higher weights overall, while 'cilos' and 'merec' generally assign lower weights. This could mean 'idocriw' gives generous importance to all criteria, viewing a wide variety as potentially impactful. 'cilos' and 'merec' methods could be more conservative or selective in their approach, or they might assign higher weights to criteria not listed in this table. It's definitely crucial, though, to possess knowledge of the specific criteria and understanding the algorithms behind each MCDA method to make accurate insights.*

**Prompt 8**: *Do the weightings across different methods suggest a consensus on the importance ranking of the criteria?*

**Response**: *The results are quite diverse across different MCDA methods, indicating a lack of consensus on the importance ranking of the criteria. Some methods give more importance to the first criterion (Bottero et al., Rodrigues et al., bwm, critic, idocriw), but others give more weight to the second (entropy, merec) or even the sixth criterion (cilos). So, the importance of the criteria largely depends on the MCDA method applied. It would be helpful to further investigate the underlying reasons and assume a weighting method according to the specific nature and requirements of the decision problem.*

ChatGPT, when operating without context, might sometimes plunge into foundational explanations of concepts that researchers might already be familiar with. For instance, as seen in Prompts 4 and 5, the responses can occasionally be rudimentary or lacking depth. Providing specific context or applying domain expertise is crucial to derive the most valuable and insightful answers.

### 4.3 Second Case Study – Correlation Analysis

We will utilize the Pearson correlation coefficient as our analytical tool to examine the interconnections between continuous values of the MCDA methods showcased in Table 7. Figure 2 provides a color-coded matrix of these interrelationships. Shades of blue, growing deeper with intensity, denote positive correlations between methods. In contrast, hues of green and yellow symbolize near-zero and negative correlations, respectively. Remarkably, the matrix underscores pronounced negative correlations existing between certain methods.

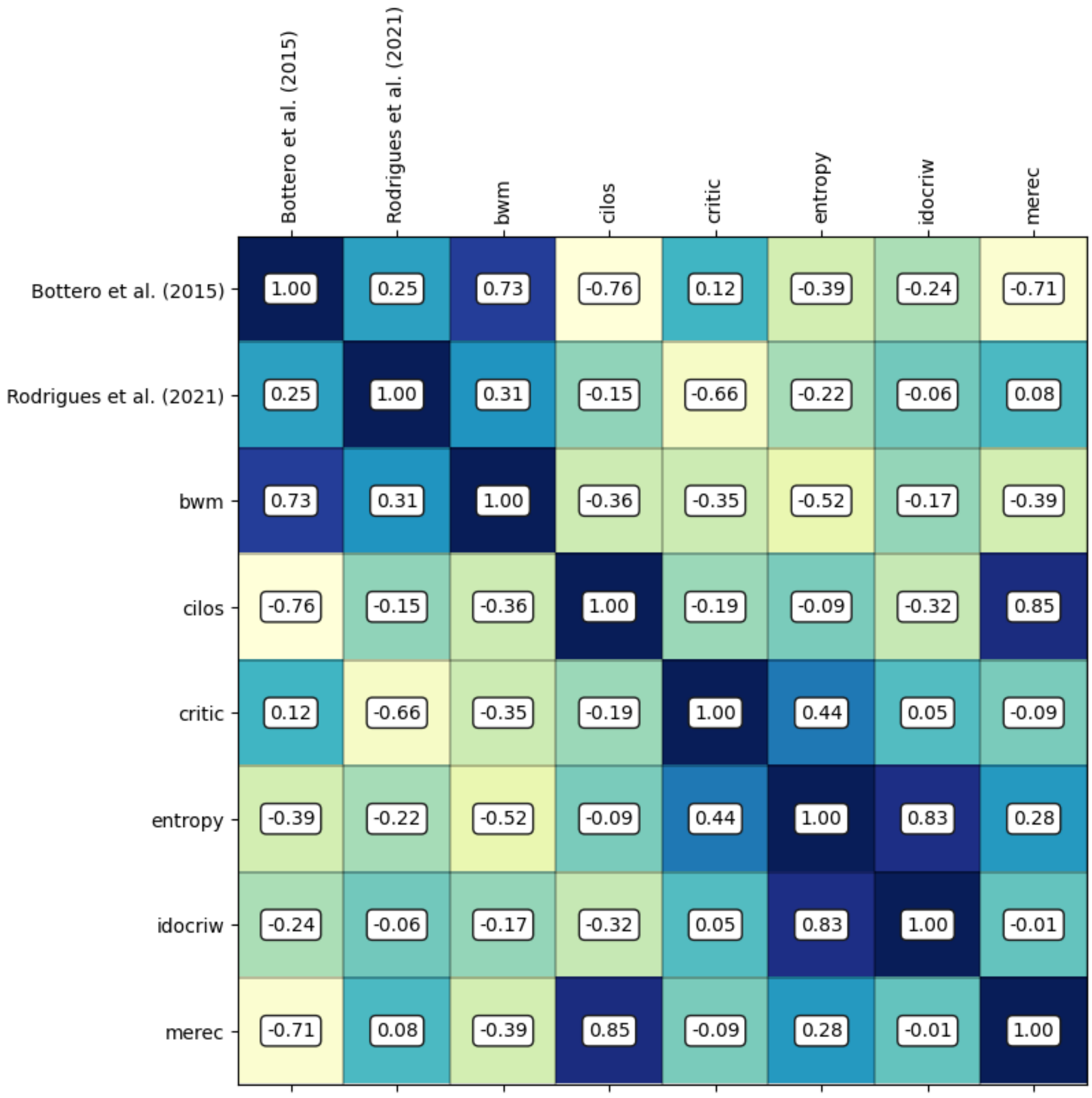

**Figure 2 – Second Case StudyWeights Correlation**

Upon examining the results from Figure 2, several standout relationships can be identified. Most prominently, CILOS and MEREC share a solid positive correlation of r = 0.85, suggesting that these methods often move in tandem. ENTROPY and IDOCRIW also display a strong positive correlation of r = 0.83, indicating a similar trend. On the other end of the spectrum, the results of Bottero et al. (2015) and CILOS exhibit a pronounced negative correlation of r = -0.76. Another significant negative relationship is observed between CRITIC and the results from Rodrigues et al. (2021), with a correlation coefficient of r = -0.66.

### 4.4 Second Case Study – Correlation Analysis – Large Language Model Aid

To extract deeper insights and potentially uncover hidden patterns in Figure 2, we'll leverage the analytical capabilities with ChatGPT (model gpt4). We plan to use a sequence of targeted prompts to augment our initial analysis by shedding light on aspects we might have overlooked.

**Prompt 1**: *Explain the significance of analyzing the correlation of weights between different MCDA methods.*

**Response**: *Analyzing the correlation of weights between different Multiple Criteria Decision Analysis (MCDA) methods is significant in several ways: 1. Checks Consistency: By examining the correlation, we can check the consistency of the decision-making process. If different MCDA methods are giving conflicting results for certain weight comparisons, it indicates potential inconsistency and warrant a re-examination of the decision problem's structure and data. 2. Validating MCDA method: This correlation data can also* *help in validating and comparing the reliability of different MCDA methods. If two methods show a high correlation of weights, they are likely to be similarly effective. In contrast, a low correlation might indicate one method is not as reliable. 3. Uncovers Biases: The correlation analysis can also uncover potential biases that might not be evident when each method is considered independently. 4. Improving Decision-Making Process: High correlation strengthens our confidence in the decision-making process, while low correlation could stimulate further investigation to uncover the potential reasons and subsequently improve the decision-making process. 5. Integrating multiple MCDA methods: If the weights' correlation between two methods is found to be high, one might think of integrating the two methods into a 'hybrid' approach which may potentially improve the robustness of the model. In the given table, the positive values suggest a positive correlation, i.e., as the weight increases in one method, it also increases in the other. On the other hand, negative values suggest a negative correlation, i.e., as the weight increases in one method, it decreases in the other method. The values close to 1 or -1 indicate strong positive or negative correlation, respectively, while those near 0 indicate weak or no correlation..* [sic]

**Prompt 2**: *What might cause differences in the weighting of criteria across various MCDA methods?*

**Response**: *The differences in the weighting of criteria between various Multi-Criteria Decision Analysis (MCDA) methods can be caused by several factors: 1. Different Interpretations of Criteria Importance: The perceived importance of a criterion can change based on who is assessing the criteria. Different people may prioritize different elements based on their perspectives. For instance, experts and laypersons, or different demographic groups, will likely perceive the importance of individual criteria differently. 2. Choosing Different Methods: Different MCDA methods use different algorithms that interpret the importance of a criterion differently. They employ distinct efficacy measures. For example, the Analytic Hierarchy Process (AHP) uses a pairwise comparison approach to defining weights, while the PROMETHEE method uses a preference function. 3. Variations in the Form and Type of Data: Different MCDA methods are better suited to different types of data (quantitative, qualitative, interval, ordinal, etc.), which can affect the weighting. 4. Sensitivity to Uncertainty: Some MCDA methods are more sensitive to uncertainty than others. If the data or preferences are uncertain, the methods may give different weights to different criteria. 5. Context and Domain of Application: The specific application or context of the decision problem might influence the weighting process. For example, in environmental decision-making, ecological criteria could be weighed more heavily than economic factors, and this could shift depending on the method. 6. Bias and Subjectivity: As MCDA reflects human judgment, personal biases and perceptions could influence the choice of criteria and the assigned weights. These can vary depending on the MCDA method being used..* [sic]

**Prompt 3**: *How can the correlation of weights between different MCDA methods impact the final decision?*

**Response**: *The correlation of weights between different Multi-Criteria Decision Analysis (MCDA) methods can greatly impact the final decision as it determines the agreement or disagreement between these methods. 1. Positive Correlation: If two MCDA methods have a high positive correlation like the Bottero et al. (2015) method and bwm method, this means that they broadly agree in the way they assign weights to the criteria. A decision made using these methods would likely be similar as they deem the same criteria to be important. 2. Negative Correlation: If two MCDA methods have a high negative correlation like the Bottero et al. (2015) method and the cilos method, this means they disagree in how they rank the importance of the criteria. A decision made using these methods could be very different which could complicate the decision-making process. 3. No Correlation: If two MCDA methods have correlation near zero (For instance, eigenvalue and idocriw method), this means there is no linear association between the weights assigned by these methods. Without additional information, we cannot deduce how a decision would vary if different combinations of these methods were used. Highly correlated MCDA methods may help to reinforce a decision if they converge towards the same outcome, thus, providing additional confidence in the decision made. However, if there is disagreement or lack of correlation between MCDA methods, this could suggest that the decision is more complex, with different methods highlighting different priorities. Therefore, the decision-makers may wish to scrutinize the differences, investigate the underlying causes and ultimately, choose an approach that best suits their problem and context. Remember that correlation* *does not imply causation - while it shows the relationship, it does not provide insights into why these weights are influencing each other in such way. It's also crucial to consider, MCDA is only one step in decision-making process, and decisions should not be made solely on the results of MCDA. Other factors, including expert opinion, regulatory considerations and public response, also need to be taken into account.* [sic]

**Prompt 4**: *What could be the implications if there is a high correlation of weights across different MCDA methods?*

**Response**: *If there is a high correlation of weights across different Multi-Criteria Decision Analysis (MCDA) methods, it means that the methods are giving almost similar and identical results. This could have several implications:  1. Consistency: In a decision-making process, consistency is key, and thus a high correlation could signify that there is a high level of consistency in the decision-making process. 2. Confirmation of validity: Strong correlation may serve as validation that the methods applied are giving accurate and valid outcomes. If similar weights are attributed to the same alternatives across different MCDA methods, it is more likely that the results are a true reflection of the situation. 3. Ease of decision making: If there is a high correlation across different MCDA methods, it means that there is agreement on which alternatives are preferred, making it easier to make a final decision. However, it is important to keep in mind: 1. Risk of Redundancy: If all methods are giving almost similar results, using different MCDA methods may be inefficient, as effort and resources may be spent on applying and analyzing multiple methods when one might have been sufficient. 2. Limits innovation: There may be instances where a decision-maker is looking for diversified analysis and results. In such cases, a high correlation of weights reduces the scope for finding innovative solutions. 3. Reliability: A high correlation might also suggest that the methods are not as reliable as they seem, as they do not provide different perspectives on the decision problem. It could indicate that the methods are being influenced by the same underlying factors rather than each providing a unique analysis. Overall, understanding these correlations can create opportunities to use the best suited MCDA method for a given context or to blend results from various methods for a more comprehensive analysis.* [sic]

**Prompt 5**: *What strategies can be used to address inconsistencies in the weights assigned by different MCDA methods?*

**Response**: *Looking at the table above, it's apparent that there are some notable differences and inconsistencies in the weights assigned by different Multi-Criteria Decision Analysis (MCDA) methods. Here are some strategies that could be used to address these inconsistencies: 1. Decision Matrix: Use a decision matrix to compare the weightings assigned by different methods. This may involve ranking the methods based on their ability to accurately reflect decision-makers' preferences. 2. Comparison of Methods: All the MCDA methods may not be equally valid for the kind of data or decision issue in hand. Carry out a comparison of the methods to understand which is the most appropriate for the specific decision context, and prioritize their weightings/refine them accordingly. 3. Average: You may consider averaging the weights determined by all methods for each criterion. This simple method is useful to address inconsistencies between different methods, although it may not always lead to the most desirable outcomes. 4. Sensitivity Analysis: Conduct a sensitivity analysis to study the effects of changes in the input data (i.e., changes in the weights assigned by different MCDA methods) on the final decision. 5. Meta-Modeling: Meta-models are a type of model that summarizes the knowledge gained from running an original model several times. This can be used for reconciliation of the inconsistencies. 6. Expert Judgement: Incorporate the judgment of the decision-makers or experts in refining the weights. It may also be helpful to consider qualitative views on the relative importance of different criteria. 7. Weighting Techniques: Use well-established weighting techniques that could better encapsulate the decision-maker's preferences. Techniques such as the Analytic Hierarchy Process (AHP) or the Best-Worst Method (BWM) could be used. 8. Cross-Check: Cross-check with real-world outcomes to see if one method's weights tend to be more accurate than the others.  Remember, these strategies are not mutually exclusive and can often be used in conjunction with one another to arrive at the best possible outcome.* [sic]

ChatGPT can occasionally misinterpret the nuances of specific techniques or methods. For example, Prompt 2 correctly described AHP's approach to weight generation but inaccurately claimed that the PROMETHEE method utilizes a preference function for the same purpose. While the model can offer valuable insights, it's essential to cross-check its responses for accuracy.

## 5. Conclusion

In conclusion, the PyDecision library, implementing over 70 MCDA methods, offers a comprehensive toolkit for researchers and practitioners to explore, analyze, and solve complex decision-making problems. Its integration into the Python ecosystem provides several advantages over traditional MCDA software. Its open-source nature invites collaboration and continuous improvement, ensuring that the library remains up-to-date with the latest advancements in the field. Furthermore, its integration with ChatGPT offers an additional layer of AI-powered analysis, although researchers should exercise caution and discernment when interpreting outputs. Notably, PyDecision has modules that facilitate comparisons between MCDA methods, whether those generating ranks or weights. Additionally, its voting methods are adapted for MCDA contexts, enabling the selection of top alternatives from rank-comparison matrices. The fusion of MCDA with AI through PyDecision modernizes both time-tested and novel approaches, exemplifying the future of decision-making tools.

## DECLARATIONS

No funding was received to assist with the preparation of this manuscript.

All authors certify that they have no affiliations with or involvement in any organization or entity with any financial interest or non-financial interest in the subject matter or materials discussed in this manuscript.

The datasets and computational details for the research presented are fully accessible to ensure transparency and reproducibility. The data for the first case study can be found at https://bit.ly/47aAVrZ, and the data for the second case study is available at https://bit.ly/3QAaJle.